\documentclass[letterpaper, 10 pt, conference]{ieeeconf}  

\IEEEoverridecommandlockouts                              

\usepackage{cite}
\usepackage{tabularx}
\usepackage{amsmath,amssymb,amsfonts}
\usepackage{graphicx}
\usepackage{textcomp}
\usepackage{color}
\usepackage{hyperref}
\usepackage[noabbrev]{cleveref}
\usepackage{multirow}
\usepackage[T1]{fontenc}
\usepackage{booktabs}

\title{\LARGE \bf
Robotic Ultrasound-Guided Femoral Artery Reconstruction of Anatomically-Representative Phantoms}

\author{Lidia Al-Zogbi$^{1}$*, Deepak Raina$^{2}$*, Vinciya Pandian$^{3}$, Thorsten Fleiter$^{4}$, and Axel Krieger$^{1}$
\thanks{*Authors contributed equally to this work.}
\thanks{This work is partially supported by the Agency for Healthcare Research and Quality (AHRQ) under Grant No. 7R18HS029124-03 (previously 5R18HS029124-03), and by the National Science Foundation (NSF) under the Future of Work at the Human-Technology Frontier (FW-HTF) Award No. 2222716.}
\thanks{$^{1}$Lidia Al-Zogbi and Axel Krieger are with the Department of Mechanical Engineering and the Laboratory of Computational Sensing and Robotics, Johns Hopkins University, Baltimore, MD, USA
  {\tt\small lalzogb1@jh.edu}.}%
\thanks{$^{2}$Deepak Raina is with the Malone Center for Engineering in Healthcare, Whiting School of Engineering, Johns Hopkins University, Baltimore, MD, USA.}%
\thanks{$^{3}$Vinciya Pandian is with the Ross and Carol Nese College of Nursing, Pennsylvania State University, University Park, PA, USA.}%
\thanks{$^{4}$Thorsten Fleiter is with the R. Cowley Shock Trauma Center, Department of Diagnostic Radiology, School of Medicine, University of Maryland, Baltimore, MD, USA.}%
}

\begin{document}

\onecolumn
\noindent This work has been accepted to the IEEE for publication. Copyright may be transferred without notice, after which this version may no longer be accessible.

\twocolumn

\maketitle

\begin{abstract}
Femoral artery access is essential for numerous clinical procedures, including diagnostic angiography, therapeutic catheterization, and emergency interventions. Despite its critical role, successful vascular access remains challenging due to anatomical variability, overlying adipose tissue, and the need for precise ultrasound (US) guidance. Needle placement errors can result in severe complications, thereby limiting the procedure to highly skilled clinicians operating in controlled hospital environments. While robotic systems have shown promise in addressing these challenges through autonomous scanning and vessel reconstruction, clinical translation remains limited due to reliance on simplified phantom models that fail to capture human anatomical complexity. In this work, we present a method for autonomous robotic US scanning of bifurcated femoral arteries, and validate it on five vascular phantoms created from real patient computed tomography (CT) data. Additionally, we introduce a video-based deep learning US segmentation network tailored for vascular imaging, enabling improved 3D arterial reconstruction. The proposed network achieves a Dice score of 89.21\% and an Intersection over Union of 80.54\% on a new vascular dataset. The reconstructed artery centerline is evaluated against ground truth CT data, showing an average $L_2$ error of 0.91$\pm$0.70 mm, with an average Hausdorff distance of 4.36$\pm$1.11mm. This study is the first to validate an autonomous robotic system for US scanning of the femoral artery on a diverse set of patient-specific phantoms, introducing a more advanced framework for evaluating robotic performance in vascular imaging and intervention.

\end{abstract}

\section{Introduction}
Femoral arterial access is integral to interventional medicine, enabling procedures such as diagnostic angiography, therapeutic catheterization, and emergency interventions. Its clinical utilization has grown substantially, with use of femoral access for transcatheter aortic valve replacement in the U.S. increasing from 57.1\% in 2011 to 95.3\% in 2019 \cite{carroll2020sts}. Despite its broad adoption, femoral access remains among the most technically challenging vascular entry sites \cite{rashid2016risk}, with a 5.9\% access failure rate reported in percutaneous aortic procedures \cite{liang2019preoperative}. The procedure's complexity is compounded by anatomical variability, overlying adipose tissue, and the required accuracy in US image interpretation. Improper needle placement increases the risk of serious complications, including retroperitoneal hemorrhage, thrombosis, and other vascular sequelae \cite{rashid2016risk, stone2012complications}. Femoral arterial access is hence currently performed only by skilled physicians in controlled hospital environments.

\begin{figure}[t!]
    \centering 
    \includegraphics[width=1\columnwidth]{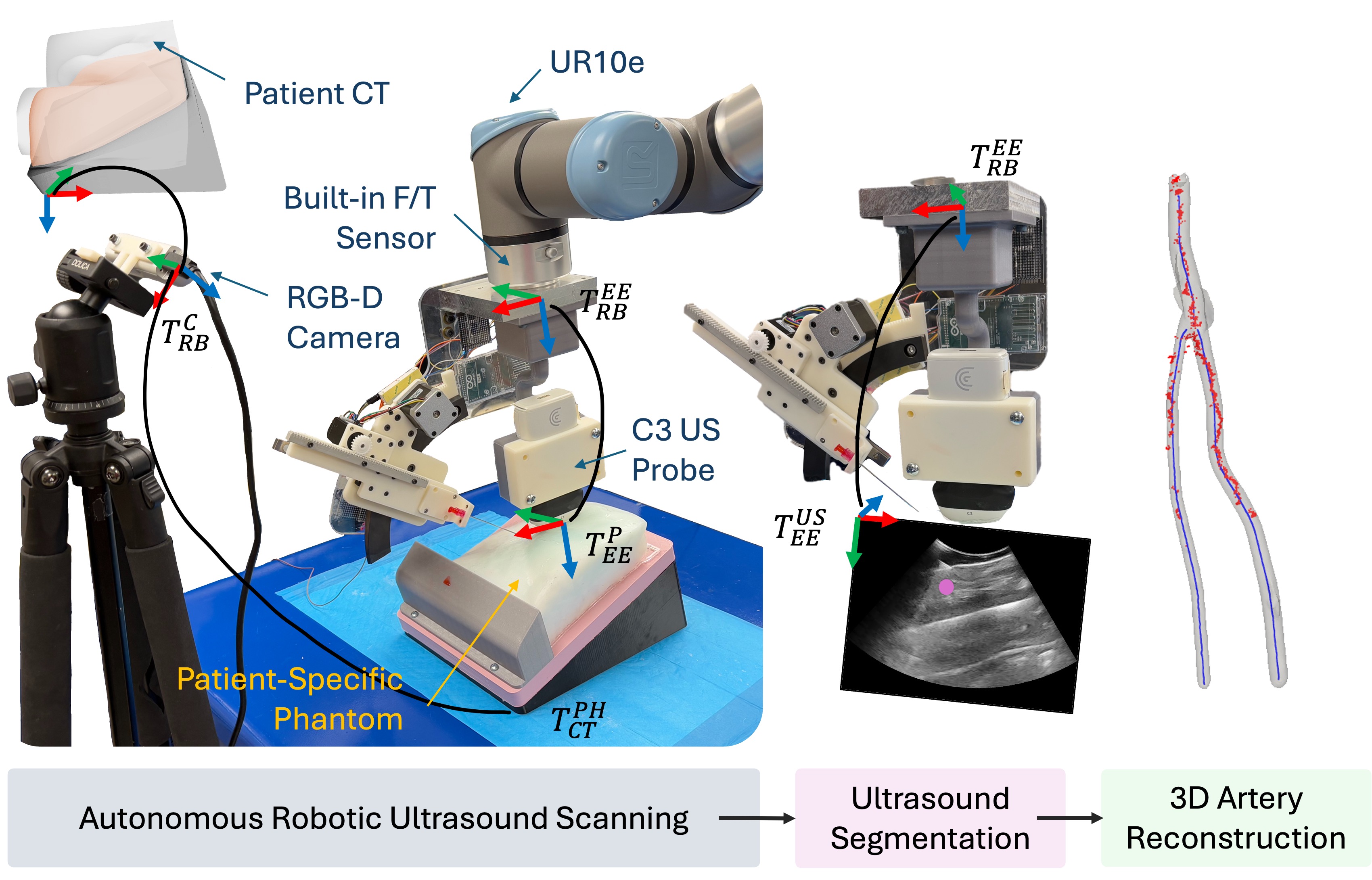}
    \caption{Overview of the robotic system and workflow for 3D artery reconstruction.}
    \label{fig:setup}
\end{figure}

Robotic systems can address aforementioned challenges by leveraging their precision and repeatability in US-guided vascular interventions. These systems can enable accurate 3D reconstruction of vasculature through automated US image segmentation, serving either as fully autonomous platforms, or as clinician-assistive technologies. Furthermore, they hold significant potential for deployment in resource-constrained environments where specialized healthcare personnel are unavailable, such as in combat zones, remote locations, or an ambulance. Although robotic US scanning and reconstruction is a mature research topic, most investigations have employed overly simplified phantom models that cannot represent the anatomical complexity encountered in clinical scenarios \cite{marahrens2022towards, ma2021novel}, or relied on single custom-designed phantoms for both system development and validation \cite{merouche2015robotic, suligoj2021robust}. To the best of the authors’ knowledge, no prior study has validated robotic femoral artery reconstruction against an independent 3D ground truth \cite{morales2024bifurcation, chen2023fully}, a critical step for reliable vascular needle placement where full arterial geometry, including bifurcation and safe entry zones, is essential for the procedure's success and patient's safety.

In this manuscript, we present an autonomous robotic system for femoral artery scanning. The proposed system integrates US image acquisition, arterial segmentation, and 3D centerline reconstruction. In particular, this work's \textbf{main contributions} are as follows: \textbf{(1)} demonstration of autonomous robotic scanning and 3D reconstruction of bifurcated femoral artery on five vascular phantoms created from real patient CT data; \textbf{(2)} development of a video-based segmentation network for axial artery segmentation in US video sequences; \textbf{(3)} comprehensive evaluation of the reconstructed bifurcated artery against ground-truth CT data. This study is the first to validate an autonomous robotic system for US scanning of the femoral artery on a diverse set of patient-specific phantoms, introducing a more rigorous framework for evaluating robotic performance in vascular imaging and intervention.

\section{Related Work}
\subsubsection{Vascular Phantoms}
Although commercial vascular phantoms provide good US image quality, their oversimplified geometries - such as rectangular prisms (CIRS model 072) or semi‑cylinders (CAE Vascular Access Phantom) - fail to replicate realistic anatomy, hindering the transfer of robotic solutions or other approaches to real clinical procedures. Larger models such as the Blue Phantom Femoral Access trainer offer improved anatomical representation, but they are often prohibitively expensive. Perhaps most significantly, these phantoms typically derive from a single anatomical model, limiting their representation of human variation, particularly for individuals with higher body mass indices (BMI). Recently, wearable training systems combined with US simulation software, such as the SonoSkin trainer (Simulab, Seattle, WA), garnered more attention. The systems consist of a skin-like cover that can be placed over a mannequin or person, paired with a simulated US probe that triggers predetermined US images at specific anatomical locations. While such systems enable basic examination practice, their reliance on pre-loaded images and lack of orientation-specific feedback limits their utility in developing image acquisition skills, or advancing US image interpretation. Additionally, creating a custom vascular phantom requires self-healing materials and durable vessels to withstand repeated needle punctures, a technology largely protected as proprietary by commercial manufacturers. We provide a simple cost-effective solution to overcome these challenges. We develop five novel patient-specific phantoms by modifying the CIRS model 072 (Sun Nuclear, U.S.) vascular phantom's surface topology and overall orientation to better represent realistic anatomy, while retaining the commercial phantom's established acoustic and mechanical properties.
\subsubsection{Ultrasound Vascular Segmentation}
Automatic segmentation of US images acquired during robotic scanning presents significant challenges, particularly when integrated with force‑control strategies. Although force control enhances safety during probe manipulation, it introduces variability in probe coupling and contact pressure. These effects, compounded by the inherently low signal‑to‑noise ratio and susceptibility to acoustic artifacts in US imaging, hinder accurate vascular segmentation.

Earlier works in US vessel segmentation used geometric and dynamic models, typically requiring seed points initialization in the first frame \cite{ma2018accurate, patwardhan20124d, mistelbauer2021semi}. Other studies have improved performance by combining Doppler and B‑mode US \cite{keil2012combining, tamimi2017automatic, moshavegh2016hybrid}.
With the emergence of deep learning, most vessel segmentation approaches employed U‑Net–based architectures that operate on individual US frames \cite{smistad2016vessel, zhou2019deep, zhou2021deep}. 
More recently, the introduction of transformers for modeling long‑range dependencies has motivated the integration of attention mechanisms into U‑Net–like models to incorporate temporal memory from US video sequences. However, to the best of author's knowledge, most studies have focused on obstetrics \cite{zhao2023ultrasound}, breast sonography \cite{chang2021weakly}, and echocardiography \cite{deng2024memsam}, with comparatively little focus on video‑based vascular US segmentation in robotic scanning.

\begin{figure}[b!]
    \centering
    \includegraphics[width=1\columnwidth]{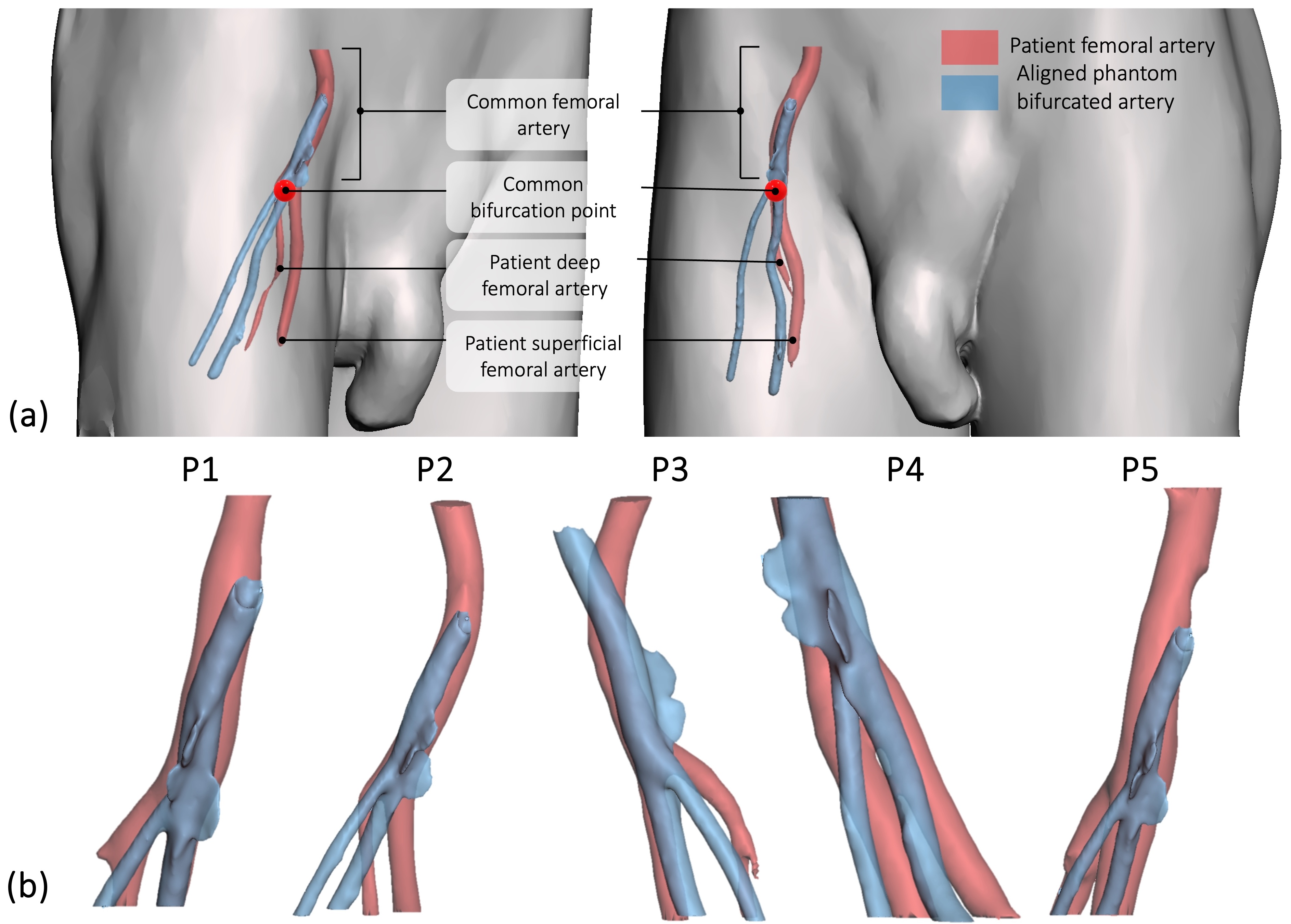}
    \caption{(a) Phantom's vessel alignment with patient's femoral artery achieved by matching the 3D position of the bifurcation point and the orientation of the common femoral artery. (b) Aligned phantom artery relative to patients' femoral arteries.}
    \label{fig:art}
\end{figure}

\subsubsection{Robotic Ultrasound Vascular Scanning} 
Robotic US has emerged as a promising solution to address several drawbacks of hand-held US imaging for abdominal \cite{al2021autonomous, raina2021comprehensive, raina2023rusopt, raina2024coaching} and cardiac \cite{amadou2024goal} procedures. In particular to vascular imaging, prior research has focused on vessel segmentation, tracking, and localization. Jiang \textit{et al.} \cite{jiang2021autonomous} used segmented vascular structures to align a robotic probe perpendicular to the anatomy and later refined spatial positioning using 3D point cloud data of the scanning surface \cite{jiang2022precise}. Chen \textit{et al.} \cite{chen2020deep} used color Doppler with B-mode images and analyzed them through deep learning for vascular access.
Merouche \textit{et al.} \cite{merouche2015robotic} conducted an automatic tracking of artery and 3D reconstruction by initializing the cross-sectional view of the artery. Ning \textit{et al.} \cite{ning2021autonomic} employed a weakly supervised US vessel segmentation network for probe orientation estimation, and a force-guided reinforcement learning agent to maintain optimal probe angle. These systems, however, were validated on either commercial or custom-designed phantoms, which fail to fully capture the anatomical complexity of real humans. While some studies have validated their approaches on animals \cite{morales2024bifurcation} or human subjects \cite{ning2021autonomic}, they were primarily limited to US image acquisition, with no independent ground truth to assess 3D reconstruction quality. In contrast, our proposed robotic US vascular scanning system is validated on patient-specific phantoms with their corresponding CT scans, providing a more representative assessment of reconstruction quality.

\begin{figure}[t!]
    \centering
    \includegraphics[width=1\columnwidth]{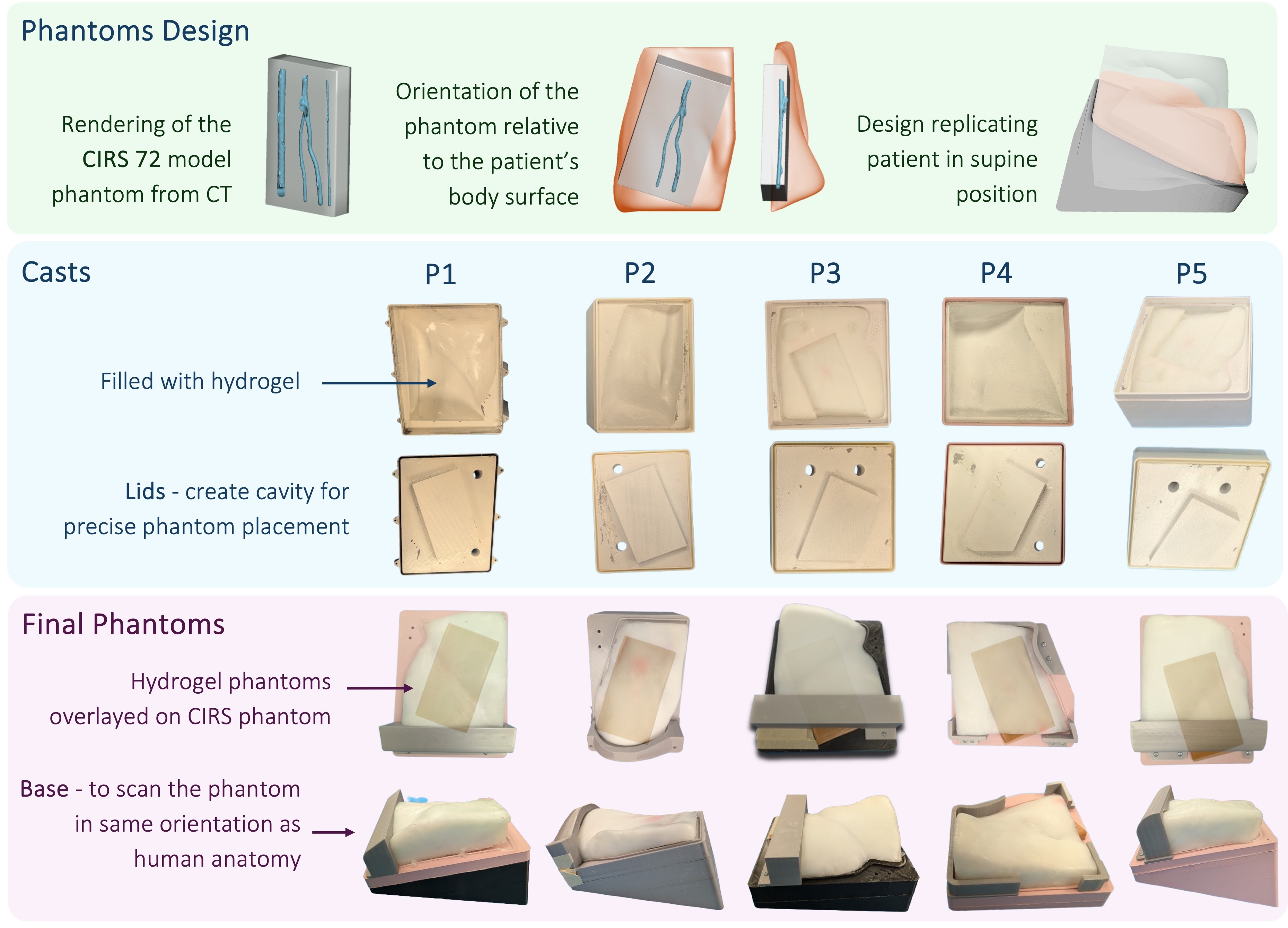}
    \caption{Visualization of the conceptual framework underlying the developed phantoms, methodology employed in their fabrication, and the final patient-specific phantoms created for all five cases.}
    \label{fig:phantoms}
\end{figure}

This manuscript is structured as follows: Section III.A describes the methodology and rationale for vascular phantom development. Section III.B details the autonomous robotic US scanning protocol and introduces a novel video‑based US segmentation network. Section IV presents and analyzes the experimental results from robotic US image acquisitions. Section V concludes the paper and outlines directions for future work.
\section{Materials and Methods}
\label{sec:matmethods}
\subsection{Patient-Specific Vascular Phantoms} \label{sec:phantoms}
The phantoms comprise a commercial model with embedded vasculature as the base, overlaid with our custom‑made hydrogel layers to more accurately replicate human anatomical surface topology. The commercial CIRS 072 model was used as the base phantom. It contains four embedded vessel structures: two lateral vessels of different diameters, and a central bifurcating vessel serving as our experimental target. To reduce detection artifacts, the blood-mimicking fluid was drained from the two lateral vessels. We utilized two CIRS phantoms: an older specimen whose hardness increased due to aging, and a newer phantom with softer material properties. Five anonymized trauma patient CT scans (four male, one female) served as references for generating phantom surface topology, with corresponding surface meshes of skin, skeleton, and vasculature. The patients had varying BMI, as increased adipose tissue commonly complicates femoral arterial access. Clinically, access targets the common femoral artery segment between the bifurcation and inguinal ligament. Unlike commercial phantoms with parallel vessels, human femoral arteries exhibit natural curvature. To enhance anatomical fidelity, we use the Meshmixer CAD software to precisely align the phantom's bifurcated vessel with patient-specific anatomy by matching the 3D position of the bifurcation point and direction of the common femoral artery (see Figure \ref{fig:art}). Although bilateral data were available for each patient, anatomical symmetry was assumed, and a single side was randomly selected for phantom generation.

Following alignment of the phantom's bifurcating vessel with the patients' femoral artery, the phantom volume is digitally subtracted from the patient’s body volume (see Figure \ref{fig:phantoms}). The resulting complementary volumes  are used to create molds for customized 10\% w/w polyvinyl alcohol hydrogel layers, which replicate patient‑specific surface topology when placed atop the commercial phantoms. Hydrogel fabrication follows a three‑day freeze/one‑day thaw cycle, after which the hydrogel layer is demolded. To achieve anatomically accurate supine positioning, support platforms are designed from patient CT data (see Figure \ref{fig:phantoms}) to orient the commercial phantom, and ensure precise alignment of the hydrogel layer with the underlying CIRS models.

\begin{figure}[b!]
    \centering
    \includegraphics[width=1\columnwidth]{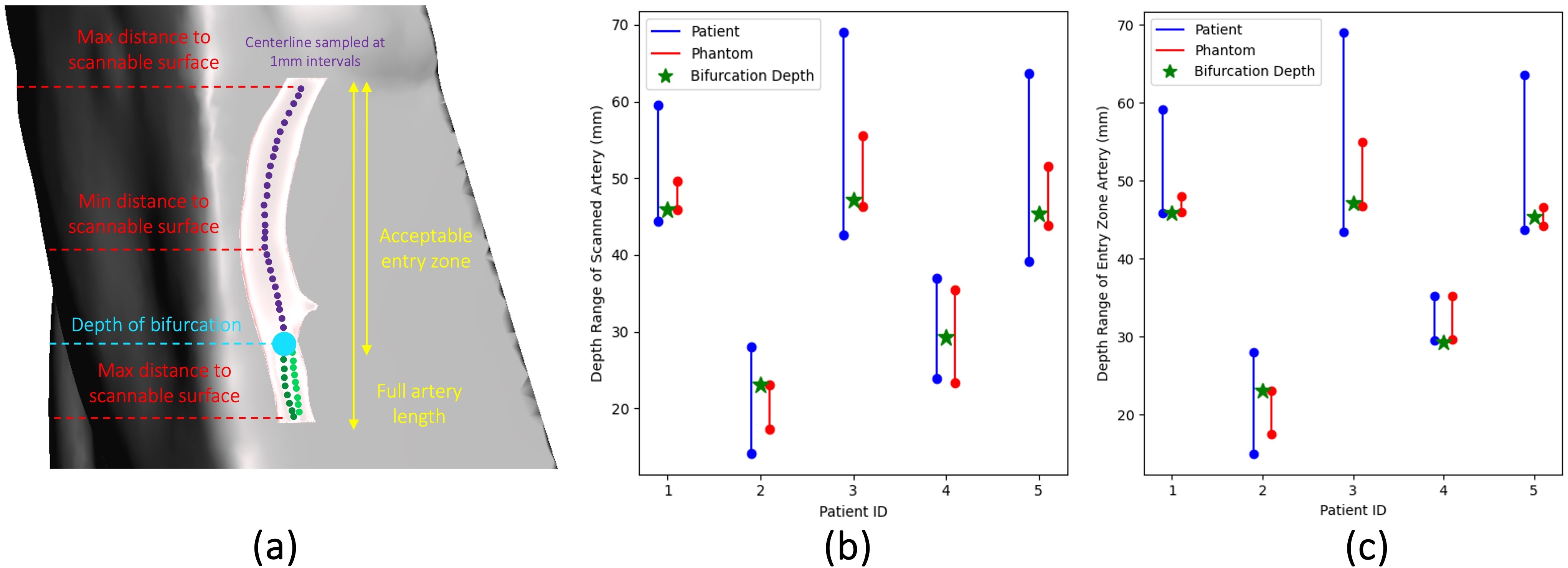}
    \caption{a) Visualization of the variables used to compute depth metrics. Comparative plots showing the depth range of (b) the entire artery that can be scanned and (c) the safe access entry zone of the artery, including the bifurcation depth, for both patients and phantoms.}
    \label{fig:metrics}
\end{figure}

Figure \ref{fig:metrics} compares the designed phantom models to patient anatomies using two metrics: the shortest $L_2$ distance from skin surface to bifurcation point, and depth ranges for both the acceptable arterial entry zone and complete artery. While patient arteries demonstrate greater depth variation due to natural curvature compared to the phantom's planar vessel, both exhibit comparable depth ranges and significant variability by design (12-56mm) across all cases.

\subsection{Femoral Artery Reconstruction}
Accurate 3D reconstruction of the femoral artery is essential for identifying critical landmarks, such as the bifurcation, that define safe needle entry points. Precise localization of the entry zone improves needle insertion planning and reduces complications, including arterial trauma and suboptimal catheter placement. This study focuses on achieving high‑fidelity femoral artery reconstruction using the methodology described below.

\subsubsection{Robotic Ultrasound System}
The robotic system, shown in Figure \ref{fig:setup}, comprises a UR10e manipulator with an integrated force/torque sensor (Universal Robots, Denmark), a convex US probe (C3 HD, Clarius Mobile Health, Canada) mounted on the robot's end effector, and a RealSense camera (Intel Corporation, U.S.) attached to a tripod. 

Given the predefined anatomical placement of the femoral artery within each phantom, the system bypasses vascular localization and instead focuses on autonomous scanning from one end of the artery to the other. In a real clinical scenario, this information can be provided to the robotic system by a paramedic. In the presented pipeline, this information is extracted using the phantom's point cloud and pre-operative CT scan. The phantom’s surface is first scanned with a RealSense camera to produce a 3D point cloud. Using the known arterial start and end points from the CT scan, a straight‑line trajectory is computed. The phantom surface is approximated by a best‑fit plane, and a perpendicular plane passing through the straight‑line trajectory is defined. The intersection of the perpendicular plane with the phantom surface point cloud defines the final US scanning trajectory, which is then discretized at 1mm intervals and supplied to the robot controller as execution waypoints. The US probe orientation is held constant along the scanning path, with its $z$‑axis aligned to the normal of the best‑fit plane (blue arrow on probe tip in Fig.\ref{fig:setup}) and its $x$‑axis (red arrow) aligned perpendicular to the initial straight‑line trajectory, thereby fully defining the probe's pose.

During scanning, the robotic system employs a force-control strategy implemented through Universal Robots' built-in force/torque control framework. This algorithm concurrently regulates the end-effector’s applied force, positional accuracy, and orientation. The scanning procedure was conducted with a contact force of 6 N for the older CIRS phantom, and 4 N for the newer model, consistent with values commonly reported in clinical practice for vascular imaging \cite{suchon2025ranges}. Throughout the scanning process, the robot’s end-effector poses and corresponding US images are recorded in a rosbag file. Synchronization is maintained using ROS2 timestamps to ensure temporal alignment between positional data and acquired US frames.

To ensure seamless integration of all system components, a unified reference frame is established, as illustrated in Figure \ref{fig:setup}. The notation $T^A_B$ denotes the homogeneous transformation that maps coordinates from frame $A$ to frame $B$. The essential reference frames employed in this system include: $T^P_{EE}$, the transformation describing the position and orientation of the US probe relative to the robot's end effector; $T^{C}_{RB}$, the transformation between the camera frame and the robot's base; $T^{US}_{EE}$, the transformation between the US image and the robot's end effector; $T^{PH}_{CT}$, the transformation aligning the physical phantom with its corresponding CAD model obtained from CT scans.

\textbf{US Image Calibration:} $T^{US}_{EE}$ is determined through a point-based calibration method \cite{zhang2018phantom}. The procedure involves acquiring US images of a spatially fixed point in a water bath from multiple robot poses, and solving $B_i X^{-1}p_i = B_j X^{-1}p_j$ using least squares method. Calibration was performed at an US imaging depth of 100mm, maintained across all experiments. It is important to note that separate calibration matrices would be required for different imaging depths due to depth-dependent scaling variations.

\textbf{Eye-to-Hand Camera Calibration:} A checkerboard was rigidly fixed to the robot table, and a calibration tool tip mounted on the robot end effector was guided to the checkerboard intersection points. The tool’s position relative to the robot base was recorded while the 3D camera captured a depth image of the checkerboard. The transformation between the camera and robot base was then estimated via least‑squares.

\textbf{Phantom-to-Patient Registration:} Scan start and end points are first defined on patient‑specific CAD models reconstructed from CT scans. To map these points to the physical phantom, the transformation between the phantom in the camera frame and its corresponding CAD model is estimated. An initial alignment is obtained by matching centroids and principal component axes, followed by refinement using the Iterative Closest Point (ICP) algorithm.

\subsubsection{Vascular Segmentation Network}
We propose a deep learning network to segment vessels in US videos by extending a traditional single-frame segmentation network into a video segmentation pipeline. The network processes preceding frames using encoders to extract key and value features, which are stored in a memory module. Since US images often exhibit speckle noise, low contrast, and transient artifacts, the memory module leverages temporal context from preceding frames to help maintain spatial continuity in the segmentation. The memory module retains the value features alongside their corresponding key features, which are subsequently accessed and integrated with query feature maps through an attention-based module. The fused features generated by this process are then passed to a decoder, which produces the final segmentation mask. An overview of the network architecture is presented in Fig. \ref{fig:network}. Key components of the network are explained below:

\begin{figure}[b!]
    \includegraphics[width=1\columnwidth]{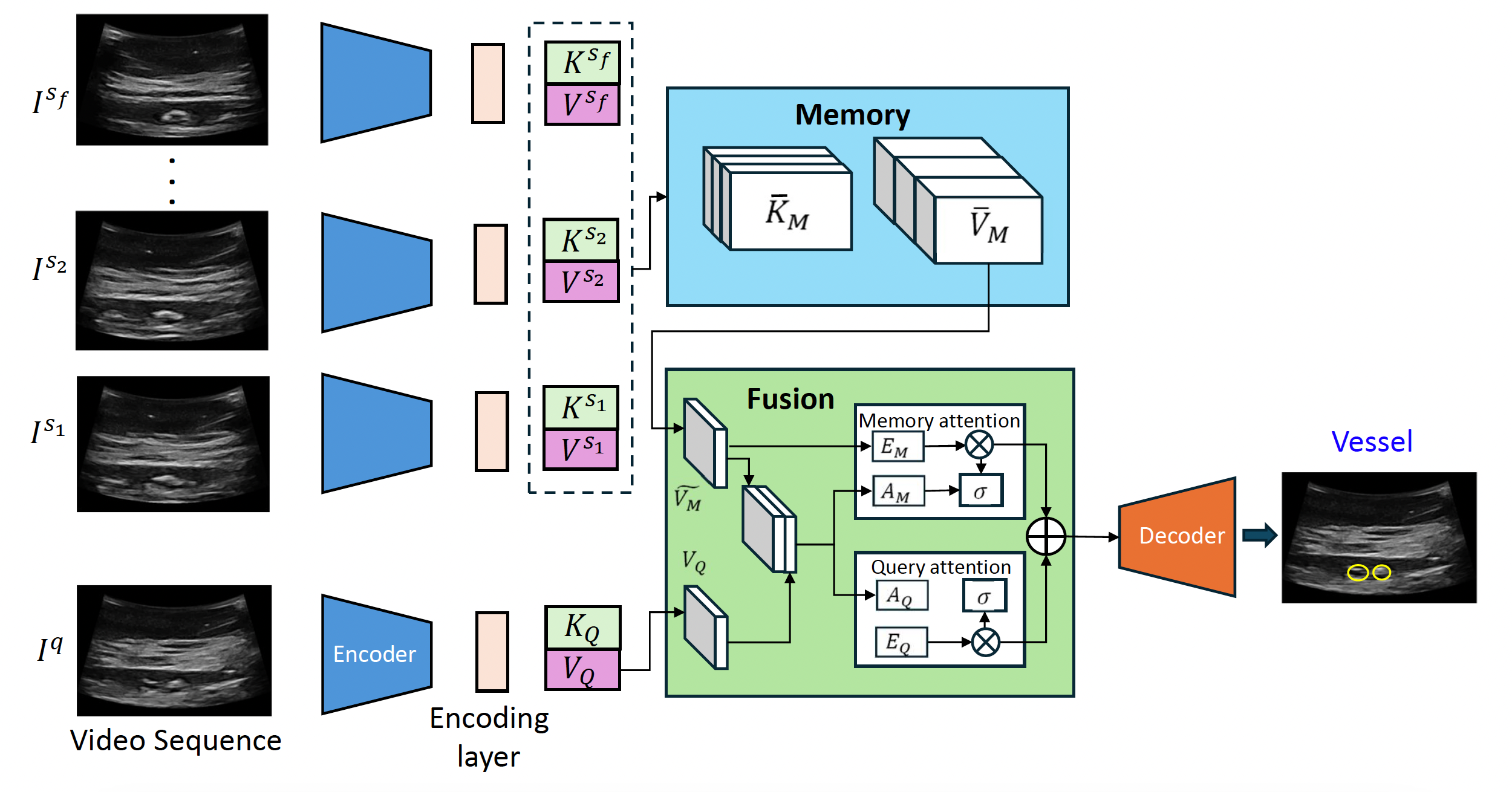}
    \caption{Overview of the vascular segmentation network.}
    \label{fig:network}
\end{figure}

    \textbf{Encoder and Decoder:} For the encoder and decoder components, we utilized the TransUNet architecture \cite{chen2021transunet}, a widely recognized network for medical image segmentation. Specifically, we selected the R50-ViT-B16 variant of TransUNet. The encoder output, initially consisting of 256 channels, was further processed through an additional encoding layer to generate key and value features with 128 and 16 channels, respectively.
    
    \textbf{Memory:} The extracted key and value features are stored in a memory module. During the segmentation of the query frame, temporal information must be retrieved from memory. This retrieval is accomplished by matching the key feature maps from the memory and the query frames, denoted as $\boldsymbol{{K}}_M^{s_i}$ and $\boldsymbol{{K}}_Q$, respectively. To optimize this process, we leverage temporal prior information as in \cite{paul2021local}, which assumes that the content in the query frame is more likely to appear in similar pixel locations in the preceding (memory) frames. This operation is implemented as a correlation operation:
    \begin{equation}
    \begin{split}
        \boldsymbol{\mathcal{X}}_{nm}^{s_i} (x,y) &= \boldsymbol{\mathcal{X}}(\boldsymbol{{K}}_M^{s_i}, \boldsymbol{{K}}_Q) \\ &= \boldsymbol{{K}}_M^{s_i}(x,y)^T \boldsymbol{{K}}_Q(x+n,y+m).
        \end{split}
    \end{equation}
    Here, $(x, y) \in [1, H] \times [1, W]$, and the pair $(n, m) \in \mathcal{R} = [-h, h] \times [-w, w]$ defines the region of search for the correlation, with $h<<H$ and $w<<W$. A softmax layer is then applied to convert the correlation maps into probability maps through
    \begin{equation}
        \boldsymbol{\mathcal{P}}_{nm}^{s_i}(x,y) = \frac{\exp(\boldsymbol{\mathcal{X}}_{nm}^{s_i}(x,y))}{\sum_{s_i \in \mathcal{S}} \sum_{(n,m) \in \mathcal{R}}\exp(\boldsymbol{\mathcal{X}}_{nm}^{s_i}(x,y))}.
    \end{equation}
    The probability maps represent attention weights, which are used to weigh both spatial and temporal information stored in the memory. The final output tensor from the memory can be expressed as
\begin{equation}
\begin{aligned}
\Tilde{\boldsymbol{{V}}}_M(x, y) &=
\sum_{s_i \in \mathcal{S}} 
\sum_{(n,m) \in \mathcal{R}}
\boldsymbol{\mathcal{P}}_{nm}^{s_i}(x,y) \\
&\quad \times \boldsymbol{{V}}_M^{s_i}(x+n,y+m),
\end{aligned}
\end{equation}
    which represents the weighted sum of all memory values $\boldsymbol{V}_M^{s_i}$ across both temporal domain and spatial domains. 

    \textbf{Fusion:} The fusion module combines $\boldsymbol{V}_Q$ and $\Tilde{\boldsymbol{{V}}}_M$ to form the decoder input needed for segmenting the query frame.
    To achieve this, an attention branch is assigned to each fusion input, as in \cite{paul2021local}. These attention branches, equipped with sigmoid-activated gates, regulate the flow of information by controlling which features - either from the query or from memory - are passed to the decoder. The fusion module can be mathematically expressed as
    \begin{equation}
    \begin{split}
        \boldsymbol{F}(\boldsymbol{{V}}_Q, \Tilde{{\boldsymbol{V}}}_M) & = \sigma(\boldsymbol{A}_Q(\boldsymbol{{V}}_Q, \boldsymbol{{V}}_M)) \otimes \boldsymbol{E}_Q(\boldsymbol{{V}}) \\ 
        & + \sigma(\boldsymbol{A}_M(\boldsymbol{{V}}_Q, \boldsymbol{{V}}_M)) \otimes \boldsymbol{E}_M(\boldsymbol{{V}}),
    \end{split}
    \end{equation}
    where $(\boldsymbol{E}_Q, \boldsymbol{A}_Q)$ and $(\boldsymbol{E}_M, \boldsymbol{A}_M)$ are the learned convolution layers of the Query attention and Memory attention, respectively. The dimension of the fusion module output is identical to that of the encoder output.

    \textbf{Dataset:} US images were acquired using the Clarius C3 HD scanner, with a hydrogel block placed over only the old CIRS phantom. The block matched the phantom’s length and width but had a variable thickness, ensuring that the acoustic propagation characteristics were consistent with those of the patient-specific phantoms. Scans were performed along all four vessels at a 100mm depth, capturing a total of 4,802 images across 10 continuous scan sequences. Vessel structures were manually segmented by two annotators using the VIA tool \cite{dutta2019via}. The dataset was partitioned into training, validation, and testing in an 80:10:10 ratio (3,838, 476, and 488 images, respectively), ensuring proportional sequence representation across all subsets. Additional dataset information is provided in Table \ref{ch5-tbl1}.

 \begin{table}[b!] 
     \caption{Overview of the acquired vascular US dataset on Phantom 1.}
        \centering 
        \resizebox{\linewidth}{!}{%
        \begin{tabular}{c|cccccccccc}
        \toprule
        Sequence ID & 1 & 2 & 3 & 4 & 5 & 6 & 7 & 8 & 9 & 10\\ 
        \midrule
        US Frames & $313$ & $453$ & $428$ & $428$ & $343$ & $594$ & $560$ & $554$ & $662$ & $467$\\
        \bottomrule
        \end{tabular}%
        }
        \label{ch5-tbl1}
\end{table}

\subsubsection{Vascular Centerline Estimation}
Following robotic US image acquisition and vessel segmentation, post‑processing is applied to suppress noise and reduce artifacts. False positive detections are eliminated by excluding centroids located beyond ±85 pixels from the image center, corresponding to a 45mm width threshold (based on a 0.2694 mm/pixel scale factor at 100mm scanning depth). This threshold is empirically derived through analysis of femoral artery characteristics across the five patient models, where the maximum observed femoral artery bifurcation width was 39.68mm bilaterally. To account for variability, a 12.5\% margin is incorporated, adjusting the threshold from 40 to 45mm. Contour detection is applied to the refined segmentation mask, retaining only contours with non-zero area. An ellipse is fitted to each valid contour to approximate vessel geometry, recording its centroid and area. Centroids are then transformed from pixel to 3D world coordinates using the US calibration matrix, with their stacked positions forming the reconstructed artery centerline. CT scans of both CIRS phantoms are acquired, and two distinct CAD models of the bifurcated vessels are generated. The vessels' centerline is then extracted using 3D Slicer and the Vascular Modeling Toolkit \cite{izzo2018vascular}, discretized at 1mm intervals to serve as the ground truth.

\subsubsection{Evaluation Metrics} To evaluate the reconstruction accuracy, the detected centerline is aligned to the corresponding ground truth using ICP. Spatial accuracy is quantified by computing the Hausdorff distance and $L_2$ norm between the reconstructed centroids and ground truth centerline. The Hausdorff distance $H(A,B)$ between two point sets $A$ and $B$ is defined as:
\begin{equation}
  H(A,B) = \max \{ \sup_{a\in A} \inf_{b\in B} ||a-b||, \sup_{a\in A} \inf_{b\in B} ||b-a||\},
\end{equation}
where $||\cdot||$ is the Euclidean distance. The Hausdorff distance in this case is computed from the reconstructed phantom centroids to the discretized ground truth centerline, since the reconstructed centerline is more sparse compared to the ground truth. The $L_2$ norm is computed as:
\begin{equation}\label{ch5-eq-l2}
  L_2 = \sqrt{\sum^N_{i=1} ||p_i - q_i||^2},
\end{equation}
where $p_i$ and $q_i$ represent the closest corresponding points on the reconstructed and ground truth centerlines, respectively, and $N$ is the total number of points.

Results are reported for both postprocessed detected centroids, and manually filtered centroids that exclude points not corresponding to the vessel. This evaluation does not assess the quality of US image segmentation per se, which is evaluated separately, but rather focuses on quantifying potential distortions in the 3D reconstruction of the artery centerline using our approach.

\begin{figure*}[t]
  \centering
  \includegraphics[width=\textwidth]{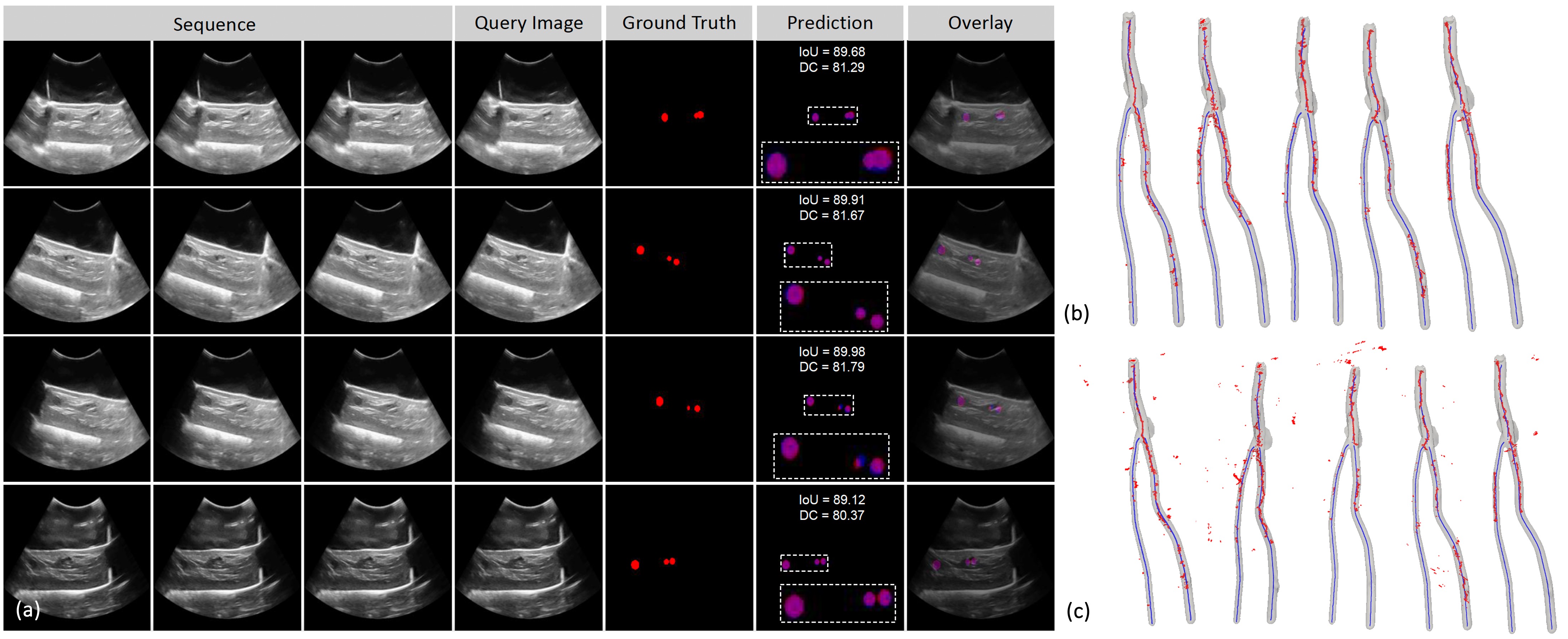}
  \caption{(a) Qualitative results of the network's segmentation on the vascular test set. Arterial 3D centerline reconstructions (red) overlaid on centerlines from the five phantoms (blue) using the older CIRS model (b) after and (c) before data filtering.}
  \label{fig:results_3D}
\end{figure*}

\section{Results and Discussion}

\subsection{Vascular Segmentation in US Images}
The network was trained on an NVIDIA GeForce RTX 4090 GPU (24GB) using Python 3.8, PyTorch 2.4, and CUDA 12.1. The Adam optimizer was initialized with a learning rate of $2 \times 10^{-4}$, adjusted by a factor of 0.8 upon plateau detection, with a minimum threshold of $1 \times 10^{-9}$. A batch size of 16 was used. The input consisted of video sequences incorporating the previous 3 frames ($s_f=$ 3). The TransUNet model was initially trained from scratch, with pretrained encoder weights subsequently transferred to the video-based model. A correlation window size of 5 and a memory sequence length of 4 were used for feature fusion. US images and corresponding ground truth labels were resized to $256 \times 256$ pixels using bilinear and nearest-neighbor interpolation, respectively. Data augmentation techniques included random horizontal flipping, contrast adjustment via color jitter, rotations within $\pm30^{\circ}$, and gamma correction with values randomly sampled between 1.0 and 2.5. The network was optimized using a joint Dice and Binary Cross-Entropy loss function for a maximum of 50 epochs.

The network's segmentation performance was assessed using the Dice Coefficient (DC) and Intersection over Union (IoU). On the training set, the model achieved a DC of 91.28\% and an IoU of 83.95\%, while on the test set, it attained a DC of 89.21\% and an IoU of 80.54\%, demonstrating good segmentation accuracy and generalization. The minimal performance disparity between training and test sets suggests that the model effectively mitigates overfitting. Figure \ref{fig:results_3D}(a) presents qualitative results from the test set, illustrating predicted segmentation masks overlaid on input images. The close correspondence between predicted and ground-truth masks further validates the model’s robustness in arterial segmentation. Due to the absence of standardized benchmark datasets for vascular segmentation, direct comparative analysis with existing methodologies presents inherent challenges. Nevertheless, for contextual reference, Morales \textit{et al.} \cite{morales2023reslicing} reported an IoU of 77.9\% on a medical phantom using their proposed technique, and an average IoU of 63.8\% on pig US data. The inference time on the same GPU used for training is 6ms per 3-frame video sequence.

\subsection{Vascular Centerline Reconstruction}
\begin{figure}[b!]
    \centering
    \includegraphics[width=\columnwidth]{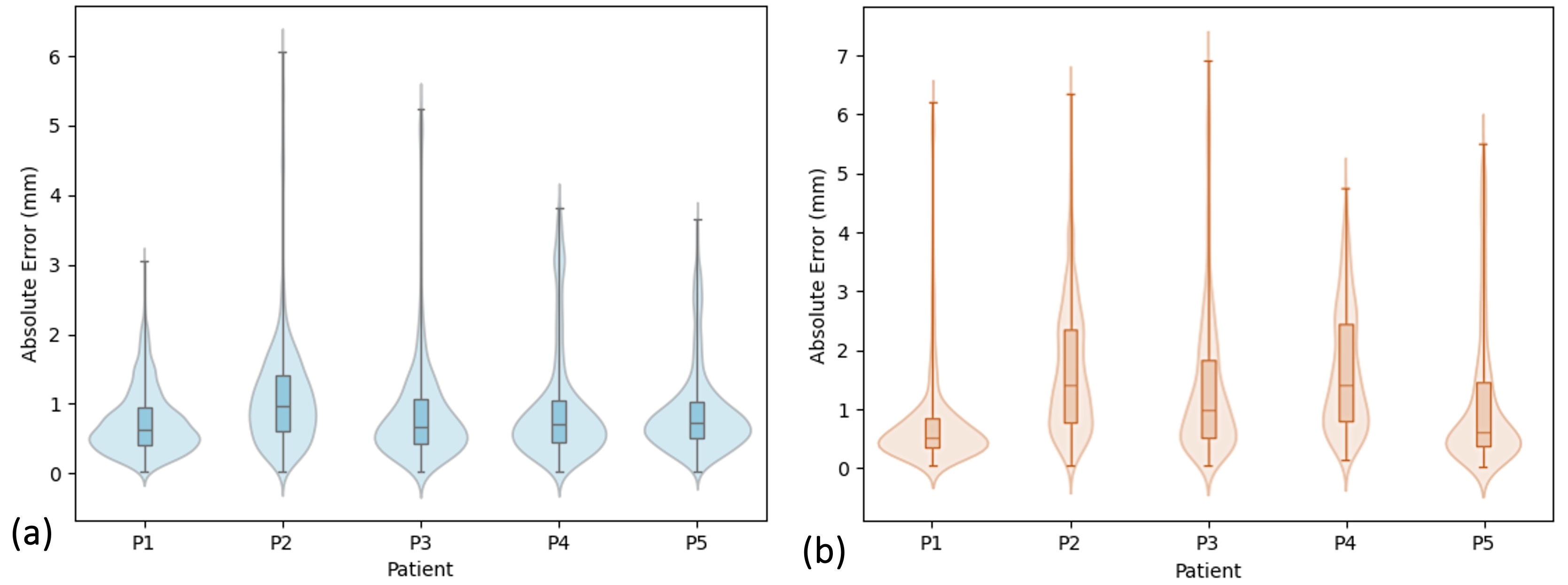}
    \caption{Violin plots of $L_2$-error in centerline reconstruction of patients (P1 through P5) for different CIRS phantoms (a) PH1 and (b) PH2.}
    \label{fig:violin}
\end{figure}

The robotic scanning protocol was completed in 1 min and 23 seconds for each of the 10 experimental cases. Examples of the reconstructed centerline overlaid on the ground truth centerline for the bifurcated arteries of the first (older) phantom are shown in Figure \ref{fig:results_3D}(b)-(c). The figure includes results both before and after manually removing outliers corresponding to erroneously detected center points by the network. Additionally, violin plots illustrating the $L_2$ errors are shown in Figure \ref{fig:violin}, with corresponding metrics such as the mean, standard deviation, and Hausdorff distance reported in Table \ref{ch5-tbl-center1}. Values in red denote the highest metrics, where values in blue denote the lowest. The number of detected centroids with and without manual filtering is also reported in Table \ref{ch5-tbl-center1}. 

\begin{table}[t!]
  \caption{Mean error, standard deviation, Hausdorff, and number of detected centroid points for all CIRS phantoms (PH1 and PH2) and patients (P1 through P5) with and without removal of outliers. All values (except for points \#) are in mm. Values in red denote highest metrics, values in blue denote the lowest.} \label{ch5-tbl-center1}
  \resizebox{\columnwidth}{!}{
  \begin{tabular}{ccc|ccc|c|}
  \cline{4-7}
 &           &      & \multicolumn{1}{c|}{\textbf{Mean $L_2$ Err.}} & \multicolumn{1}{c|}{\textbf{Std}} & \textbf{Hausdorff} & \textbf{Points \#} \\ \hline
  \multicolumn{1}{|c|}{\multirow{10}{*}{\textbf{PH 1}}} & \multicolumn{1}{c|}{\multirow{2}{*}{\textbf{P1}}} & \textbf{Unfiltered} & \multicolumn{1}{c|}{2.957} & \multicolumn{1}{c|}{6.410} & 41.77 & 2659 \\ \cline{3-7} 
  \multicolumn{1}{|c|}{} & \multicolumn{1}{c|}{} & \textbf{Filtered} & \multicolumn{1}{c|}{\textcolor{blue}{\textbf{0.741}}} & \multicolumn{1}{c|}{0.463} & \textcolor{blue}{\textbf{3.046}} & 2340 \\ \cline{2-7} 
  \multicolumn{1}{|c|}{} & \multicolumn{1}{c|}{\multirow{2}{*}{\textbf{P2}}} & \textbf{Unfiltered} & \multicolumn{1}{c|}{3.503} & \multicolumn{1}{c|}{7.613} & \textcolor{red}{\textbf{58.67}} & 2591 \\ \cline{3-7} 
  \multicolumn{1}{|c|}{} & \multicolumn{1}{c|}{} & \textbf{Filtered} & \multicolumn{1}{c|}{1.105} & \multicolumn{1}{c|}{0.788} & 6.059 & 2305 \\ \cline{2-7} 
  \multicolumn{1}{|c|}{} & \multicolumn{1}{c|}{\multirow{2}{*}{\textbf{P3}}} & \textbf{Unfiltered} & \multicolumn{1}{c|}{\textcolor{red}{\textbf{3.819}}} & \multicolumn{1}{c|}{7.026} & 31.96 & 2110 \\ \cline{3-7} 
  \multicolumn{1}{|c|}{} & \multicolumn{1}{c|}{} & \textbf{Filtered} & \multicolumn{1}{c|}{0.890} & \multicolumn{1}{c|}{0.815} & 5.241 & 1719 \\ \cline{2-7} 
  \multicolumn{1}{|c|}{} & \multicolumn{1}{c|}{\multirow{2}{*}{\textbf{P4}}} & \textbf{Unfiltered} & \multicolumn{1}{c|}{1.711} & \multicolumn{1}{c|}{4.777} & 49.56 & 2632 \\ \cline{3-7} 
  \multicolumn{1}{|c|}{} & \multicolumn{1}{c|}{} & \textbf{Filtered} & \multicolumn{1}{c|}{0.955} & \multicolumn{1}{c|}{0.821} & 3.820 & 2482 \\ \cline{2-7} 
  \multicolumn{1}{|c|}{} & \multicolumn{1}{c|}{\multirow{2}{*}{\textbf{P5}}} & \textbf{Unfiltered} & \multicolumn{1}{c|}{1.223} & \multicolumn{1}{c|}{2.449} & 21.71 & 3339 \\ \cline{3-7} 
  \multicolumn{1}{|c|}{} & \multicolumn{1}{c|}{} & \textbf{Filtered} & \multicolumn{1}{c|}{0.887} & \multicolumn{1}{c|}{0.629} & 3.647 & 3271 \\ \cline{2-7} 
  \multicolumn{1}{|c|}{} & \multicolumn{1}{c|}{\multirow{2}{*}{\textbf{Avg}}} & \textbf{Unfiltered} & \multicolumn{1}{c|}{2.642} & \multicolumn{1}{c|}{5.655} & 40.73 & 2666 \\ \cline{3-7} 
  \multicolumn{1}{|c|}{} & \multicolumn{1}{c|}{} & \textbf{Filtered} & \multicolumn{1}{c|}{0.915} & \multicolumn{1}{c|}{0.703} & 4.362 & 2423  \\ \hline
\hline
  \multicolumn{1}{|c|}{\multirow{10}{*}{\textbf{PH 2}}} & \multicolumn{1}{c|}{\multirow{2}{*}{\textbf{P1}}} & \textbf{Unfiltered} & \multicolumn{1}{c|}{2.471} & \multicolumn{1}{c|}{4.950} & 15.10 & 2045 \\ \cline{3-7} 
  \multicolumn{1}{|c|}{} & \multicolumn{1}{c|}{} & \textbf{Filtered} & \multicolumn{1}{c|}{\textcolor{blue}{\textbf{0.746}}} & \multicolumn{1}{c|}{0.836} & 6.216 & 1770 \\ \cline{2-7} 
  \multicolumn{1}{|c|}{} & \multicolumn{1}{c|}{\multirow{2}{*}{\textbf{P2}}} & \textbf{Unfiltered} & \multicolumn{1}{c|}{4.933} & \multicolumn{1}{c|}{7.923} & 19.28 & 2017 \\ \cline{3-7} 
  \multicolumn{1}{|c|}{} & \multicolumn{1}{c|}{} & \textbf{Filtered} & \multicolumn{1}{c|}{1.629} & \multicolumn{1}{c|}{1.038} & 6.355 & 1678 \\ \cline{2-7} 
  \multicolumn{1}{|c|}{} & \multicolumn{1}{c|}{\multirow{2}{*}{\textbf{P3}}} & \textbf{Unfiltered} & \multicolumn{1}{c|}{2.358} & \multicolumn{1}{c|}{4.349} & \textcolor{red}{\textbf{37.68}} & 2491 \\ \cline{3-7} 
  \multicolumn{1}{|c|}{} & \multicolumn{1}{c|}{} & \textbf{Filtered} & \multicolumn{1}{c|}{1.343} & \multicolumn{1}{c|}{1.183} & 6.922 & 2324 \\ \cline{2-7} 
  \multicolumn{1}{|c|}{} & \multicolumn{1}{c|}{\multirow{2}{*}{\textbf{P4}}} & \textbf{Unfiltered} & \multicolumn{1}{c|}{\textcolor{red}{\textbf{6.587}}} & \multicolumn{1}{c|}{6.297} & 34.71 & 1168 \\ \cline{3-7} 
  \multicolumn{1}{|c|}{} & \multicolumn{1}{c|}{} & \textbf{Filtered} & \multicolumn{1}{c|}{1.645} & \multicolumn{1}{c|}{1.035} & \textcolor{blue}{\textbf{4.743}} & 957 \\ \cline{2-7} 
  \multicolumn{1}{|c|}{} & \multicolumn{1}{c|}{\multirow{2}{*}{\textbf{P5}}} & \textbf{Unfiltered} & \multicolumn{1}{c|}{1.533} & \multicolumn{1}{c|}{2.907} & 22.72 & 2724 \\ \cline{3-7} 
  \multicolumn{1}{|c|}{} & \multicolumn{1}{c|}{} & \textbf{Filtered} & \multicolumn{1}{c|}{1.151} & \multicolumn{1}{c|}{1.249} & 5.499 & 2667 \\ \cline{2-7} 
  \multicolumn{1}{|c|}{} & \multicolumn{1}{c|}{\multirow{2}{*}{\textbf{Avg}}} & \textbf{Unfiltered} & \multicolumn{1}{c|}{3.576} & \multicolumn{1}{c|}{5.285} & 25.89 & 2089 \\ \cline{3-7} 
  \multicolumn{1}{|c|}{} & \multicolumn{1}{c|}{} & \textbf{Filtered} & \multicolumn{1}{c|}{1.302} & \multicolumn{1}{c|}{1.068} & 5.947 & 1879 \\ \hline
  \end{tabular}}
\end{table}

The 3D points reconstructed from the segmented axial US images show strong spatial alignment with the ground truth centerlines extracted from CT scans. Similar consistency was observed in results obtained from the second CIRS model, demonstrating the robustness of our proposed method. These findings demonstrate that the transformation from the segmented US space to the 3D world frame preserves spatial accuracy without introducing distortions - a critical requirement for reconstructing small‑scale vascular structures, such as the femoral artery, with diameters as small as 4mm.

The importance of filtering out erroneous centroids is best illustrated in Figure \ref{fig:results_3D}(b)-(c). The unfiltered data exhibits a naturally wider spread of points than the filtered data, making it difficult to reliably assess reconstruction quality in isolation. The Hausdorff errors are understandably larger than the $L_2$ errors, as the Hausdorff metric in this context reports the largest $L_2$ error within the dataset (as reflected by the maximum values in the violin plots). The average filtered $L_2$ and Hausdorff errors for the first vascular phantom (0.915 and 4.362mm, respectively) are smaller than those for the second phantom (1.302 and 5.947mm), as shown in Table \ref{ch5-tbl-center1}. This discrepancy likely arises from the lower quality of the US images for the second phantom, which increased the occurrence of false positives in the segmentation. It is important to highlight that the network was trained exclusively on the first CIRS model. While both phantoms are theoretically identical, their appearance under US differs in practice due to material variations and imaging artifacts. Additionally, the smaller bifurcated vessel in the second phantom, with its reduced diameter, was frequently missed by the network due to limited visibility - a challenge further exacerbated in higher‑BMI patient phantoms - resulting in sparser centroid detections for that vessel. The use of a convex US probe may have further contributed to this issue. Probe contact with the phantom did not always ensure full transducer-surface coupling, leading to suboptimal imaging. This effect was particularly pronounced in scenarios where the natural body contours would necessitate different probe orientations, even for a skilled clinician. A linear probe, commonly used for vascular access procedures, could improve acoustic coupling and, in turn, enhance segmentation performance.

Accurate US segmentation alone does not guarantee realistic 3D reconstruction; however, our approach achieves high-fidelity reconstructions even in complex phantoms with irregular topology, including arteries located as deep as 69mm in patients with higher BMI.

\section{Conclusions and Future Work}
While robotic US for vascular access has seen significant advancements, prior studies have not validated vessel 3D reconstruction on anatomically representative phantoms. In this work, we present a robotic US-guided system for femoral artery reconstruction on a diverse set of 5 patient-specific phantoms. Our approach achieves high-fidelity vessel reconstructions even in complex phantoms with irregular topology, including deeper arteries in patients with higher BMI. Additionally, we adapted a video-based segmentation network for axial artery segmentation in US video sequences. The reconstructed artery centerline exhibited strong agreement with patient-specific CT data, the clinical gold standard for vascular imaging. These results represent a step forward in enhancing robotic US-assisted vascular access procedures, strengthening their potential for clinical translation.

Future work will focus on extending this framework to evaluate the robotic system’s capability for fully autonomous vascular needle insertion, including incorporating autonomous identification of the initial and final scanning points, and identification of a safe needle entry point from the reconstructed bifurcated artery. Comprehensive validation studies will also be conducted, including experiments on animal models and fresh cadaver specimens, to rigorously evaluate the system’s performance in realistic clinical environments.


\bibliography{bibliography}
\bibliographystyle{ieeetr}

\end{document}